\documentclass{article}
\usepackage{ijcai19}
\usepackage{fancyhdr}
\usepackage{times}
\usepackage{soul}
\usepackage{url}
\usepackage[hidelinks]{hyperref}
\usepackage[utf8]{inputenc}
\usepackage[small]{caption}
\usepackage{graphicx}
\usepackage{amsmath}
\usepackage{booktabs}
\usepackage{algorithm}
\usepackage{algorithmic}
\usepackage{latexsym}
\usepackage{color}
\usepackage{url}
\usepackage{algorithm}
\usepackage{algorithmic}
\usepackage{graphicx}
\usepackage{xspace}
\usepackage{makecell}
\usepackage{subcaption}

\usepackage{enumitem}
\usepackage{url}
\usepackage{graphicx} 
\usepackage{fixltx2e}
\usepackage{mathtools}
\usepackage{multirow}
\usepackage{mathtools}

\usepackage{algorithm}
\usepackage{algorithmic}

\usepackage{amsmath}
\usepackage{amsfonts}
\usepackage{amssymb}
\usepackage{array}

\DeclareMathOperator*{\argmin}{arg\,min}
\DeclareMathOperator*{\argmax}{arg\,max}

\newcommand{\spoke}{\xspace{\it Spoke}\xspace}
\newcommand{\ignore}[1]{}

\newcommand{\statics}{\ensuremath{s_{static}}}
\newcommand{\adas}{\ensuremath{s_{adapt}}}
\newcommand{\sumtopk}{\ensuremath{\mathbf{sum\text{-}top\text{-}k}}}

\newcolumntype{?}{!{\vrule width 1pt}}

\usepackage{booktabs} 

\fancypagestyle{plain}{ %
  \fancyhf{} 
  
\cfoot{\small Proceedings of the 2019 IJCAI Workshop SCAI: The 4th International Workshop on Search-Oriented Conversational AI}
}

\begin{document}
\thispagestyle{plain}

\title{Practical User Feedback-driven Internal Search Using Online Learning to Rank}

\author{
Rajhans Samdani, Pierre Rappolt, Ankit Goyal, Pratyus Patnaik\\
\{rajhans, pierre, ankit, pratyus\}@askspoke.com
}

\maketitle

\begin{abstract}
We present a system, \spoke, for creating and searching internal knowledge base (KB) articles for organizations. \spoke is available as a SaaS (Software-as-a-Service) product deployed across hundreds of organizations with a diverse set of domains. \spoke continually improves search quality using conversational user feedback which allows it to provide better search experience than standard information retrieval systems without encoding any explicit domain knowledge. We achieve this by using a real-time online learning-to-rank (L2R) algorithm that automatically customizes relevance scoring for each organization deploying \spoke by using a query similarity kernel. 

The focus of this paper is on incorporating practical considerations into our relevance scoring function and algorithm that make \spoke easy to deploy and suitable for handling events that naturally happen over the life-cycle of any KB deployment. We show that \spoke outperforms competitive baselines by up to 41\% in offline F1 comparisons.
\end{abstract}

\vspace{-0.1in}
\section{Introduction}
This paper presents our system, \spoke, for storing and searching Knowledge Base (KB) articles for different organizations. \spoke is available as a SaaS (Software-as-a-Service) product that can be used by any organization for documenting and searching over their internal workplace articles. We start by discussing salient aspects of the problem of KB management as a SaaS product (KB SaaS).

\subsection{Knowledge Base Search as a Service}\label{sec:challenges}
Each organization using \spoke\footnote{\url{www.askspoke.com}} for KB management creates a private corpus containing articles that are available only to the users from that organization. A user can query \spoke with their workplace queries, and the goal of \spoke is to respond with the right article if such an article already exists inside the KB. Table~\ref{kb-examples} shows four common domains, sample questions from these domains, and titles of KB articles (body of the article omitted for brevity) that answer these questions.

\begin{figure*}[t]
\footnotesize
\begin{subfigure}{.5\textwidth}
  \centering
  \includegraphics[width=0.61\textwidth]{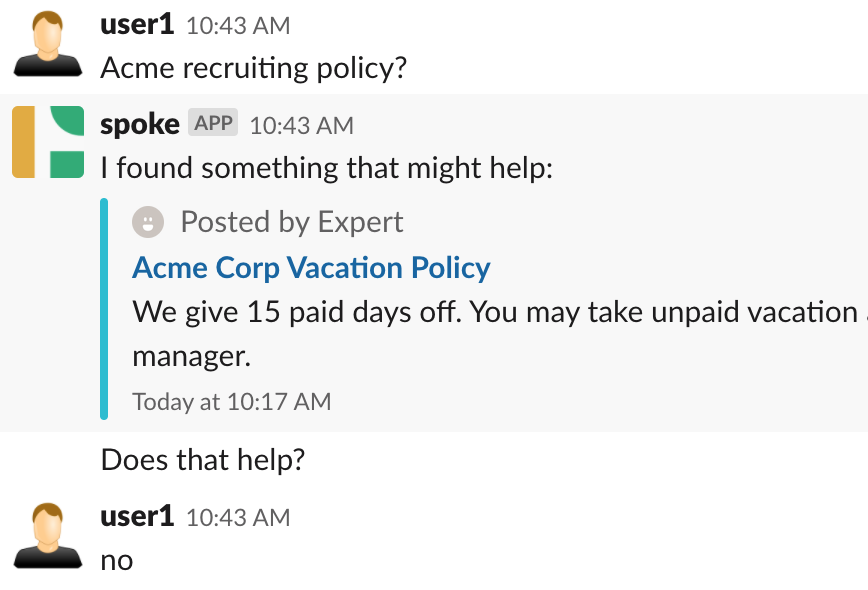}
  \caption{User1 gives negative feedback to incorrect answer.}
  \label{fig:negative_feedback}
\end{subfigure}%
\begin{subfigure}{.5\textwidth}
  \centering
 \includegraphics[width=0.61\textwidth]{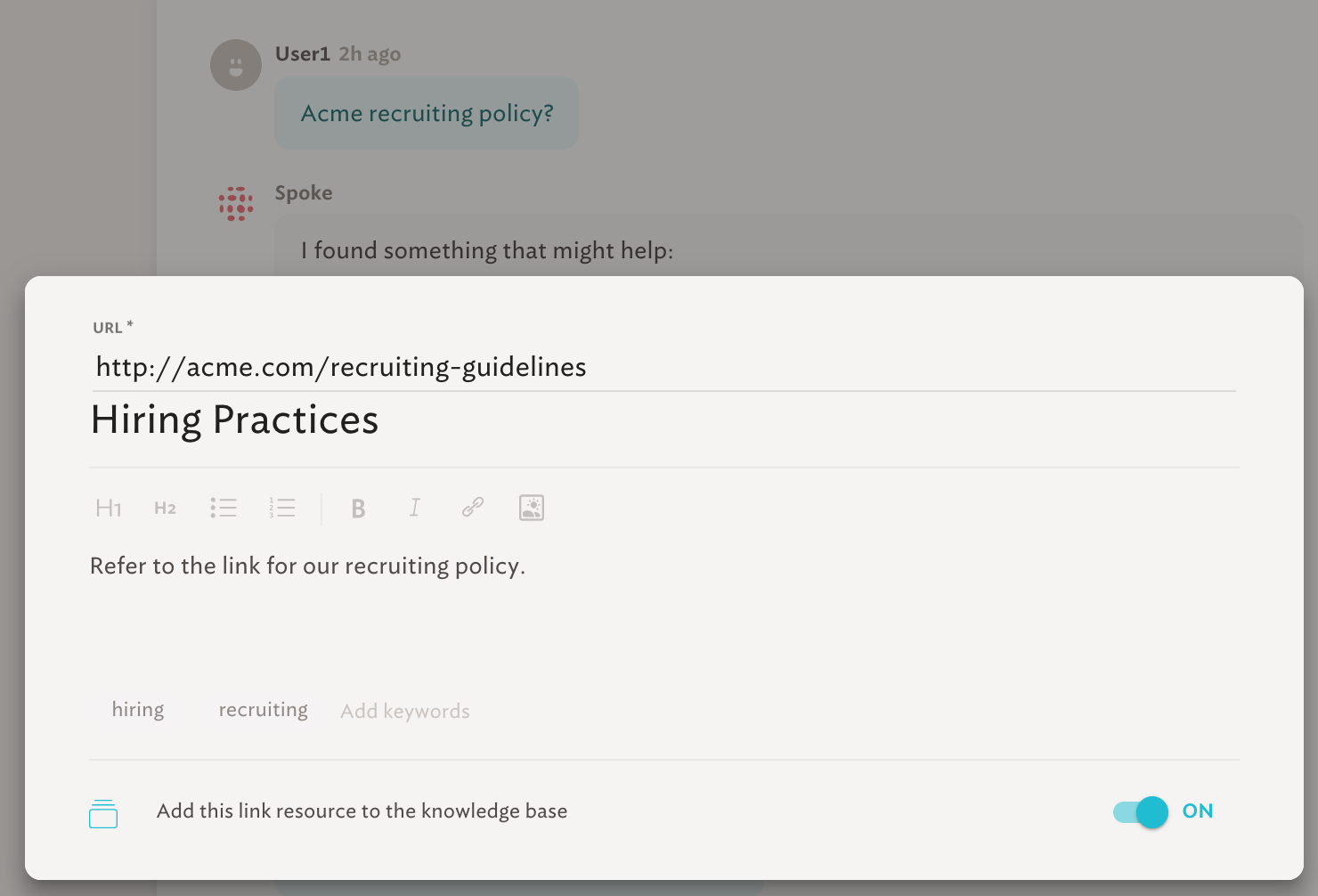}
  \caption{Expert creating a new KB article.}
  \label{fig:expert_creates_kb}
\end{subfigure} \\
\vspace{0.2in}
\begin{subfigure}{.5\textwidth}
  \centering
\includegraphics[width=0.46\textwidth]{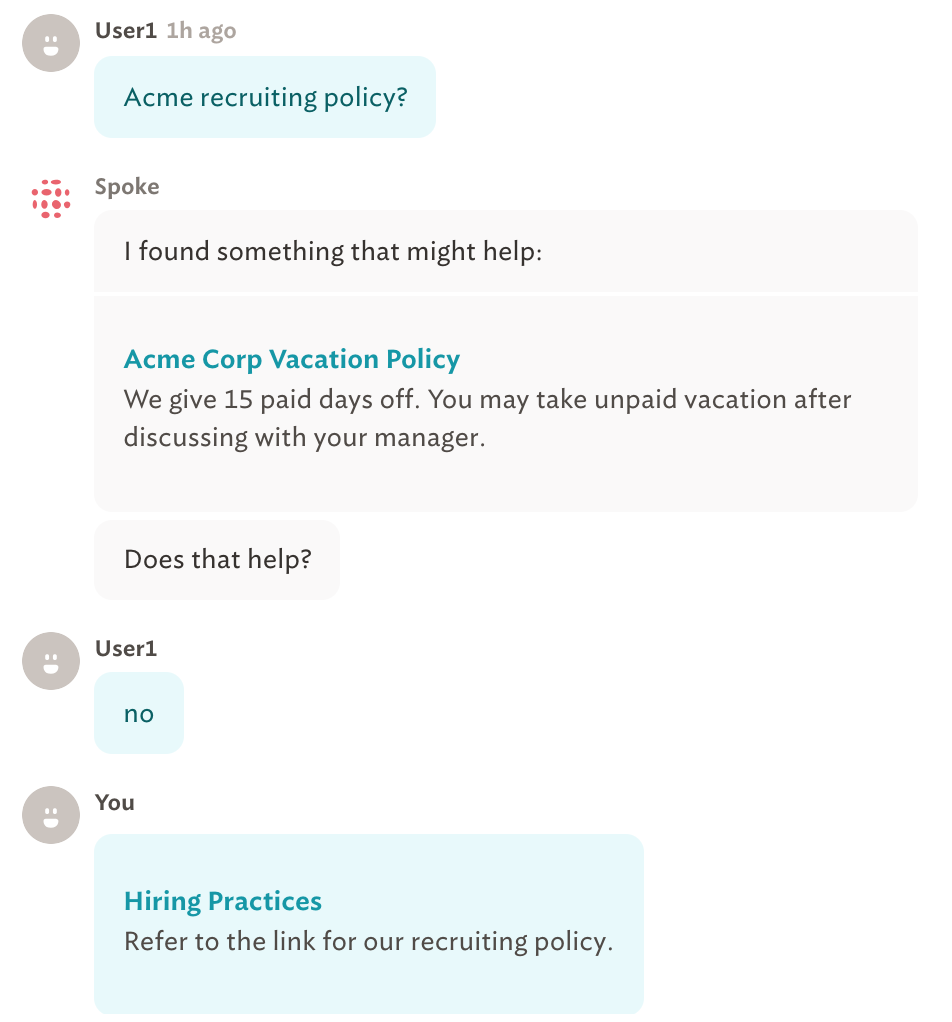} 
  \caption{Expert responding to user1 with the right KB article.}
  \label{fig:expert_feedback}
\end{subfigure}%
\begin{subfigure}{.5\textwidth}
  \centering
 \includegraphics[width=0.65\textwidth]{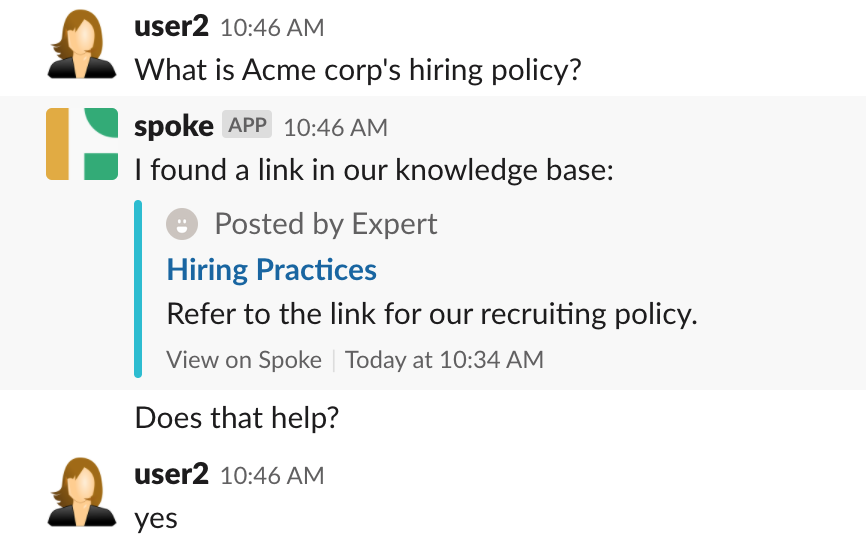} 
  \caption{User2 gets a correct answer from \spoke.}
  \label{fig:positive_feedback}
\end{subfigure} 
\caption{Example illustrating how \spoke deployed for a hypothetical organization \textit{Acme corp} learns from user and expert feedback.~\ref{fig:negative_feedback} shows a user asking a query that \spoke answers with an incorrect KB (chosen due to large term overlap) and the user rejects the answer.~\ref{fig:expert_creates_kb} shows an expert recognizing an information gap is creating an appropriate KB article in \spoke's web UI.~\ref{fig:expert_feedback} shows that user1's question is routed to the expert who is responding with the correct KB article.~\ref{fig:positive_feedback}: another user asks a similar query to \spoke but this time \spoke gives the right response. \spoke learns from the negative user feedback and the positive expert feedback and adapts.}
\label{fig:spokeux} 
\normalsize
\end{figure*}

\begin{table*}[h]
\small
\begin{center}
\begin{tabular}{|c|c|c|}
\hline KB Domains & Sample Knowledge-seeking Questions & Sample KB titles  \\ \hline
\makecell{Information Technology (IT)} & \makecell{``How do I get on the VPN?" \\ ``My macbook froze. Help!"} & \makecell{``Connecting to the VPN" \\ ``Troubleshooting Macbook"} \\\hline
\makecell{Human Resources (HR)} & \makecell{``Do we support 401k?" \\ ``Where is our recruiting rubrik?"} & \makecell{``Retirement benefits" \\ ``Hiring guideines"}  \\\hline
\makecell{Sales} & \makecell{``Maximum amount I can spend on a client dinner" \\ ``Where can I find the Q4 sales numbers?"} & \makecell{``Client Dinner Expenses", \\``Sales Dashboards"}  \\\hline 
Marketing & ``Brand assets" & ``Where is our brand logo?"\\\hline
\end{tabular}
\end{center}
\caption{\label{kb-examples} KB domains, questions for these domains sampled from our usage data, and titles of KB that answer these questions.}
\normalsize
\end{table*}

Indexing and searching over documents has been studied extensively in the information retrieval literature~\cite{ManningRaSc08,Ha96}. However, searching over internal KB is uniquely challenging when compared to web search and document retrieval tasks studied in academia~\cite{RWJHK96} for the following reasons:
\begin{itemize}[itemsep=0pt,topsep=3pt,leftmargin=*]
\item \textbf{Dynamic KB:} Real world KB are dynamic. During the lifetime of a KB deployment, new KB articles are created or existing articles may be modified; old articles get less relevant or entirely outdated with the creation of new articles. E.g. an article on \textit{Sales Process Outline} created in 2017 will become outdated in 2019 as the sales process changes. This poses a challenge for ML-based search system which must be designed to quickly unlearn old behavior with new conflicting information.
\item \textbf{Siloed Datasets:} The KBs from different organizations are siloed and so information and signals across them cannot be directly combined to train an ML system. This is unlike web search~\cite{YHTDZOCKDNLC16} where millions of query url pairs are available.
\item \textbf{Article Type:} Articles inside an organization can belong to arbitrary domains each having its specific semantics and jargon. E.g. articles from IT domains often contain names of internal servers or printers that are not a part of common knowledge. It is not scalable to inject knowledge and semantics specific to each domain in the search system. Furthermore, articles can take various forms, which may not be easily indexed: files (pdfs, Microsoft Word Docs, etc), images, hyperlinks, etc. 
\item \textbf{Scalable ML:} \spoke is deployed in thousands of separate organizations, thus it is not possible to separately train an ML-based search model for each deployment. 
\end{itemize}
\ignore{
Machine learning (ML) techniques such as learning-to-rank (L2R)~\cite{Burges10,Joachims02} have also been extensively studied for web search~\cite{YHTDZOCKDNLC16}.
We improve the quality of ML-driven information retrieval against all these challenges by exploiting user feedback in a conversational search setup.
}
\subsubsection{Limitations of Internal KB Search over Web Search}
In the last few decades, the web search experience has improved significantly using multiple signals e.g. graph-based signals~\cite{Kleinberg99,BrinPa98}, anchor text~\cite{ChakrabartiDoKuRaRaToGiKl99}, web click mining~\cite{JoGrPaHeGa05}, etc. However, these signals are not available for internal KB search which has been restricted to term-match based features.  While neural networks based approaches~\cite{GyselRiKa18,ChNaDo17,BWGCSQCW10} show great promise at learning relevance and semantics, they require a large amount of in-domain data, which is not possible in KB SaaS setup due to as the datasets are siloed. To this end, we have designed our system \spoke to extract more signal by using user interactions.

\subsubsection{Feedback-driven search experience in \spoke}\label{sec:system}
We show how organizations typically use \spoke and illustrate the feedback-driven search process.  Organizations usually have a few experts who are in charge of helping end users with their questions. These experts are also responsible for creating KB articles in \spoke to answer user questions.
Each KB article in \spoke has a four user-supplied fields that can be indexed: \textsc{title}, \textsc{body}, \textsc{keywords}, and \textsc{link}. Users  issue their questions via conversational media like chat. \spoke  responds with one answer or no answer (when it is not confident in the relevance of any article.)
In case a user expresses unhappiness with the results, \spoke reaches out to the experts that can then respond to the query by either creating a new KB article in response (recognizing an information gap) or by responding with an existing KB answer that \spoke missed. This process is illustrated in Figure~\ref{fig:spokeux}. Since prediction of user happiness is not the focus of this work, we  simply predict user happiness using a simple regular expressions-based system.

\subsection{Contribution}
This paper discusses the design of our KB system \spoke that address the challenges of KB SaaS discussed in Sec.~\ref{sec:challenges}. Our paper makes the following contributions:
\begin{enumerate}[itemsep=0pt,topsep=3pt,leftmargin=*]
\item Support real-time online learning to rank i.e. \spoke learns from user and expert feedback in real time. 
\item We use a novel trick to change the scoring function which allows \textit{unlearning} of old information using a constant amount of new user feedback. This allows the KB search to evolve with the updation and deletion of old KB articles and addition of new KB articles.
\item We present a relevance scoring function that explicitly models high-dimensional lexical features (e.g. raw words) in a kernelized form using query similarity functions.
\item We show that our adaptive system outperforms a strong L2R baseline by upto 41\% in offline experiments. Our system is deployed for hundreds of orgs and is continually getting better at returning relevant results. 
\end{enumerate}

\vspace{-0.1in}
\section{Relevance Scoring with Online L2R}
In this section, we will show how we design a relevance scoring function for KB SaaS  addressing the challenges listed in Sec.~\ref{sec:challenges}. We also present an algorithmic overview of KB management in \spoke.

\subsection{Formal Problem Definition}
Let $q$ be a query and let $d$ be a KB article. As detailed above, we allow users to provide positive or negative feedback for a query $q$ and document $d$. Let us assume at time $t$, we get feedback $y_t \in \{-, +\}$ for query $q_t$ and document $d$. For document $d$, let us define all the positive feedback queries as:
\begin{equation}
Q^+_d = \left\{(q_t) ~s.t.~ y_t = + \right\}_{t} \enspace.
\end{equation}
Similarly define $Q_d^-$. Let $Q_d = Q_d^+ \cup Q_d^-$. Our goal is to design a system that can learn from $\mathcal{Q} = \{Q_d,~\forall d\}$ to improve relevance scoring for the organization.

\subsection{Scoring with Pairwise and Lexical Scores}
As a first step, we model the relevance score as a sum of pairwise match-based score and lexicalized score as:
\begin{align}
\nonumber  s(q, d) &=& f^1(\psi_{pairwise}(q,d); \theta_1) + f^2(\psi_{lex}(q,d); \theta_2^d) \\
&= \hspace*{-0.5in}  & \hspace*{-4in} \statics(q,d) + \adas(q,d) \enspace, \label{eq:sumscore} 
\end{align}
for appropriate functions $f^1$ and $f^2$ and features $\psi_{pairwise}$ and $\psi_{lex}(q, d)$, which are defined below.
\begin{enumerate}
\item \textit{Pairwise match features, $\psi_{pairwise}(q,d)$:} These features compute the match between query and document using different textual match-based feature extractors e.g. term-based similarity like BM25~\cite{RWJHK96}; semantic similarity like Word2Vec~\cite{MSCCD13} based dot product; synonyms match, etc. $\psi_{pairwise}(q,d)$ are computed by applying these feature extractors over the query and different textual fields of $d$ like \textsc{title} and \textsc{body}. We use around 50 match features in our system (see Sec.~\ref{sec:comparison}).
\item \textit{Lexical features, $\psi_{lex}(q, d)$:} Refers to raw words or word-based features (e.g. embeddings) that are extracted from queries in the training corpus $\mathcal{Q}$ and are associated with documents $d$. These features allow us to extract associations of specific query words with documents. For example a document about \texttt{Tax Forms} may have words like \texttt{W2}, \texttt{1099}, \texttt{IRS} associated with it as lexical features. Lexical features have large dimensionality and hence are vastly more expressive than the match features and crucial to expressing semantics of the domain in our scoring function. 
\end{enumerate}
Fuethermore, we use the pairwise match features in a static scoring function $\statics$ that is fixed for all articles (and all organizations using \spoke) and the lexical features in an adaptive scoring function $\adas$ which is trained from query feedback $Q_d$ for each organization separately. The advantage of this approach is that we can create $\statics$ \textit{offline} using pre-labeled training examples using state-of-the-art Learning-To-Rank (L2R)  techniques~\cite{Joachims02,Burges10} with only a few hundred examples while allowing customizing the overall score for each organization. We will compare our adaptive algorithm to purely static baseline in Sec.~\ref{sec:experiments}. Next we describe how we create the adaptive part of the scoring function.

\subsection{Adapt Lexical Match from Feedback}\label{sec:adpativescore}
The query feedback-based score of each document is expressed using lexical features, $\psi_{lex}$ as
\begin{equation} \label{eq:adaptivescoreprimal}
\adas(q, d) = f^2(\psi_{lex}(q,d); \theta_2^d) \enspace.
\end{equation}
Letting $\Theta = \{\theta_2^d\}$ be the set of parameters over all documents, $\Theta$ can be trained by empirical risk minimization (ERM) over examples $Q$:
\small
\begin{equation} \label{eq:erm}
\argmin_{\Theta} \hspace{-0.1in} \sum_{(q, d), y \in Q} \hspace{-0.1in} L\left(y,  f^2(\psi_{lex}(q,d); \theta_2^d) + \statics(q,d) \right) + \Omega(\Theta), 
\end{equation}
\normalsize
where $L$ is a loss function and $\Omega$ is a regularizer (e.g. l2 norm). This setup of directly training weights over lexical or word-based features is exemplified by~\cite{RadlinskiJo05,BWGCSQCW10}. 

However, rather than representing parameters over lexical features, we express the score in a \textit{dual} kernelized~\cite{LSSCW02} form\footnote{In the SVM literature, often referred to as the Kernel Trick.}:
\small
\begin{align}
\nonumber \adas(q,d)  = &\beta g \left( \{w_d(q') s_{qsim}(q, q')\}_{q' \in Q_d^+} \right) \\
- &\gamma g \left( \{w_d(q'')  s_{qsim}(q, q'')\}_{q'' \in Q_d^-} \right) \enspace, \label{eq:adaptivescore} 
\end{align}
\normalsize
where $s_{qsim}$ is an appropriate kernel function representing similarity between two queries, $w_d(q') \geq 0$ is the weight of query $q'$ for document $d$, $g: \mathbb{R}^+ \times \mathbb{R}^+ \times \hdots \Rightarrow \mathbb{R}^+$ is a function that aggregates the query similarity scores, and $\beta, \gamma \geq 0$ are constants. We justify this choice in Sec.~\ref{sec:dual}, where we show how the kernelized representation in Eq.~\ref{eq:adaptivescore} is as expressive as the primal featurized representation in Eq.~\ref{eq:adaptivescoreprimal} when $g = \textbf{sum}$ function, and in addition provides several practical advantages.

Combining~\ref{eq:sumscore} and~\ref{eq:adaptivescore}, the  relevance score for KB article $d$ is 
\small
\begin{align}
\nonumber s(q,d) =  f(\psi_{pairwise}(q,d)) +& \beta g \left( \{w_d(q') s_{qsim}(q, q')\}_{q' \in Q_d^+} \right) \\
   -  \hspace{1.0in}   &  \hspace{-1.0in} \gamma g \left( \{w_d(q'')  s_{qsim}(q, q'')\}_{q'' \in Q_d^-} \right)\enspace. \label{eq:score} 
\end{align}
\normalsize
Next, we discuss how we select the key parameters in Eq.~\ref{eq:score}.

%

\subsection{Choosing Parameters for Adaptive Scoring}
In this section, we show how we select the key parameters of $s$: the query similarity kernel $s_{qsim}$, the query score aggregator function $g$, and weights of past queries $w_d$  in Eq.~\ref{eq:score}. The hyperparameters $\beta$ and $ \gamma$ are tuned on development data.

\subsubsection{Choosing the Query Similarity Function $s_{qsim}$}
The query similarity function $s_{qsim}$ computes how similar two queries are in their intent. It is a fixed function and is constant across all \spoke deployments. This function can be a kernel function like cosine similarity over Bag-of-Words (BOW) but can also be more powerful learned functions like neural networks~\cite{BoSaBaZa15}. In our setup, we pick a simple yet expressive function, $s_{qsim} = CosineSimilarity$ with TFIDF representations over unigrams and bigrams.


\subsubsection{Choosing the Aggregation Function $g$}\label{subsec:aggregation}
As mentioned in Sec.~\ref{sec:adpativescore} (and described further in detail in Sec.~\ref{sec:dual}), if we want to mimic empirical risk minimization, we can choose $g$ to be the $\mathbf{sum}$ function. However, from a practical KB design perspective, the function $g$ must satisfy certain constraints which we discuss below.
\begin{enumerate}[itemsep=0pt,topsep=3pt,leftmargin=*]
\item \textbf{Montonocity:} $g$ should monotonically increase with larger values to ensure that replacing less similar queries with more similar increases the score:
\[ g(d') \geq g(d), \forall d' \geq d, |d'| = |d| \enspace.\]
\item \textbf{Increasing:} $g$ should monotonically increase with more positive values to ensures that adding an irrelevant query does not dilute the total query similarity output:
\[ g(d \cup \{x\}) \geq g(d),~\forall x \geq 0 \enspace.\] 
\item \textbf{Bounded Magnitude:} Assuming $s_{qsim}$ is bounded, $g$ must have a bounded magnitude:
\[ |s_{qsim}| \leq l \rightarrow |g| \leq c l \enspace,\]
for some constant $c$. This somewhat surprising constraint is motivated by practical KB maintenance concerns. As discussed in Sec.~\ref{sec:challenges}, KBs change dynamically over time, as organizations add new KB articles to their \spoke deployment and as old articles get less relevant. We want \spoke to be able to learn to give higher scorer to new articles than old (potentially outdated) articles with a bounded number of mistakes (i.e. feedbacks). Eq.~\ref{eq:sumscore} shows that each article initially has a fixed static score to which the adaptive score $\adas$ is added over time. Thus we can guarantee that new articles can get higher score than old articles with bounded mistakes only if the magnitude of $\adas$ is bounded. Boundedness of $\adas$ can be guaranteed (Eq.~\ref{eq:adaptivescore}) only if $g$ is also bounded.

\ignore{
Assume this new KB article $d^{new}$ is about \textsc{Private Network}. Now, let us say there is an old KB article $d^{old}$ with title \textsc{VPN Client}, which now may be outdated but it not deleted or updated (it is very common in organizations to not clean up old information). When user searches with query $q$ = \texttt{How do I get on company's privatenetwork?}, \spoke should respond with $d^{new}$ and not $d^{old}$.
}
\end{enumerate}

The boundedness constraint rules out $g = \mathbf{sum}$ function (suggested by the ERM approach)  since $\mathbf{sum}$ can grow unboundedly. Another reasonable aggregator \textbf{average} does not satisfy constraint 2. We propose to use $g = \sumtopk$ that computes the sum of $k$ highest values from a set wherein, for each new query $q$, we are summing the score of $k$ most similar queries from past positive (negative) queries $Q_d^+$ ($Q_d^-$), This function is bounded by $k|s_{qsim}|$ and it satisfies all of the above constraints.


\subsubsection{Training weights $w_d$ of queries}
For implementing the function in Eq.~\ref{eq:score}, we store the queries from the user and expert feedback and train their weights using an online learning algorithm. We choose online learning over more common batch training~\cite{GuoFaAiCr16,GyselRiKa18,BWGCSQCW10} as learning instantaneously is important for our product to show it's utility and win user trust. We store at most $m = 100$ positive and negative queries for each article for scalability. We adopt a version of Perceptron-style additive update algorithm implemented in the dual space as described by~\cite{ShalevSi07} which amounts to constant updates to query weights (initialized to zero). However, we make the update weights different for expert feedback ($\delta_e$) than user feedback ($\delta_u$) with $\delta_e > \delta_u$ based on a practical insight that the relevance opinion of experts is more valuable than the opinion of a user. 

\ignore{

Common online L2R algorithms use exploration-exploitation algorithm like Thompson Sampling~\cite{GrotovRi16}. However, we pick a simple mistake-driven algorithm for two reasons: 1) we show only one result to the user and cannot afford to inject noise in the result for better UX, 2) since we get experts to respond with the correct answer for a subset of queries, we do not have to completely rely on bandit feedback. 
}

\subsubsection{Algorithm}
Algorithm~\ref{al:overview} presents the overall strategy for handling various events --- KB creation, searching, and feedback --- in \spoke. Note that $\mathbb{I}$ represents the indicator function.

\ignore{
\textbf{Approximations for speed:} We make two approximations to achieve fast inference time in our algorithm:
\begin{enumerate}[itemsep=0pt,topsep=3pt,leftmargin=*]
\item Rather than ranking all KB articles for a query, we obtain a set of top 20 candidate KB articles using a fast Lucene-based retrieval system like Elasticsearch\footnote{http://elastic.co} which we then rerank using our more intensive and accurate score $s$. Furthermore, in our Lucene index, we also associate the stored queries in $Q_d^+$ with the document $d$.
\item Inference time of query similarity-based kernelized form scales with the number of past queries. We bound this time by only storing at most $m$ ($\approx 100$) most recent queries in $Q_d^+$ (or $Q_d^-$). As discussed in Sec~\ref{sec:challenges}, due to low query volume in the workplace domain, $m=100$ works well for our case.
\end{enumerate}
}

\begin{algorithm}[t]
 \small
\begin{algorithmic}
\STATE{\textbf{Input:}} $k$ (number of queries to sum), $\beta$, $\gamma$ (Eq~\ref{eq:score}), threshold $\tau$ for KB confidence, maximum number of queries to store $m$, weights for user ($\delta_u$) and expert feedback ($\delta_e$) \vspace{0.05in}
\STATE{\textbf{Before \spoke is deployed:}} Train model $f$ offline over features $\psi_{pairwise}$ to create static score $\statics$, and deploy it across all organizations using \spoke\vspace{0.05in}
\STATE{\textbf{For a \spoke deployment in an organization:}}
\FOR{time $t=1...$}
\IF{new doc $d$ created}
\STATE Index $d$; initialize $Q_d^+ = Q_d^- = \emptyset$
\ELSIF{query $q_t$ is searched}
\STATE $D \leftarrow $ candidate-KB-articles$(q_t)$
 \FOR{$d \in D$} 
 \STATE  Compute  $s_{qsim} (q, q_t), \forall q \in Q_d$ \\[-1mm]
 \STATE Pick top $k$ highest scoring  queries in $Q_d^+$ and $Q_d^-$\\[-1mm]
 \STATE Compute $s(q_t, d)$ as in Eq~\ref{eq:score}\\[-1mm]
\ENDFOR
\STATE $d_{max}, s_{max} \leftarrow \argmax(s(q_t, d)), \max(s(q_t, d))$
\STATE If $s_{max} > \tau$, return $d_{max}$ to the user

\ELSIF{received feedback $y \in \{-, +\}$ from person $u$ for $q_t, d$}

\STATE $\delta \leftarrow \mathbb{I}_{u = expert} \mathbb{I}_{y=+} \delta_e - \mathbb{I}_{u = user} \mathbb{I}_{y=-} \delta_u$
\IF{$q_t \in Q_d^y$ (query already exists in corresponding set)}
\STATE $w_d^y(q_t) \leftarrow w_d^y(q_t) + \delta$ (update the weight)
\ELSE
\STATE $Q_d^y \leftarrow Q_d^y \cup \{q_t\}, ~~ w_d^y(q_t) \leftarrow \delta$  (add new query, initializing weight)
\ENDIF
\STATE Clip $Q_d^y$ keeping only the most recently updated $m$ queries

\ENDIF 

\ENDFOR

\end{algorithmic}
\normalsize
\caption{Overview of \spoke deployed in an organization.}
\label{al:overview}
\end{algorithm}


\subsection{Why Choose the Kernelized Form?}\label{sec:dual}
We make two arguments for choosing the kernelized form (Eq.~\ref{eq:adaptivescore}) over the featurized form (Eq.~\ref{eq:adaptivescoreprimal}). 

\subsubsection{Expressiveness of the kernelized approach:} 
The kernelized scoring function in Eq.~\ref{eq:adaptivescore} subsumes the score in Eq.~\ref{eq:adaptivescoreprimal} for common feature functions $\psi$ with appropriate choice of $s_{qsim}, g$ and the constants. The reasoning follows from the celebrated Representer Theorem over Reproducing Kernel Hilbert Spaces (RKHS)~\cite{HoScSm08} (a generalization of the kernel trick). Using this theorem, we can assert that having $\psi_{lex}$ as normalized BOW features $\psi_{lex}(q, d) = tf(q) / \| tf(q) \|$ in Eq.~\ref{eq:adaptivescoreprimal} is equivalent to setting $s_{qsim}$ to CosineSim, $g$ to the $\mathbf{sum}$ function, and by setting $w_d$, $\beta$, and $\gamma$ appropriately in Eq.~\ref{eq:adaptivescore}.

\subsubsection{Practical Advantages of the kernelized approach:}
Choosing the kernelized form for implementing $\adas$ using query similarity rather than explicitly storing query features with documents confers three practical advantages. 
\begin{enumerate}[itemsep=0pt,topsep=3pt,leftmargin=*]
\item We can leverage advances in deep learning by using more expressive deep query similarity model trained offline (e.g. ~\cite{BoSaBaZa15}). Training deep models in online learning is far more challenging.
\item It gives an explicit handle over the influence of past queries (and the relative score of old documents vs new documents) by controlling the aggregation function $g$. As detailed in Section~\ref{subsec:aggregation}, we choose $g=\sumtopk$ instead of the $\mathbf{sum}$. 
\item  It allows faster deployment of new query features. E.g. consider the scenario where we change the lexical features by adding Word2Vec features to existing BOW features. In the Kernelized approach, we can achieve this without any retraining simply by replacing old query similarity function $s_{qsim} = CosineSim$ with a new query similarity function $s_{qsim} = CosineSim+ Word2VecSimiliarity$. In the explicitly featurized approach, we will have to retrain and replace adapt models for all client organizations. 
\end{enumerate}

\ignore{
\textbf{Approximations for fast response time:} To achieve fast response time, we obtain a set of top 20 candidate KB articles using a fast retrieval system like Elasticsearch\footnote{http://elastic.co} which we then rerank using $s$. One downside of the dual form is that inference time scales with the number of past queries. We bound this time by only storing at most $m$ ($\approx 100$) queries. As discussed in Sec~\ref{sec:challenges} we have relatively low query volume for each organization; $m=100$ works well for our case.
}

\ignore{
The need to perform online learning also implies choosing precludes the use of deep learning techniques for IR~\cite{GuoFaAiCr16, HHGDAH13} which 
}

\vspace{-0.1in}
\section{Related Work}
There is a rich body of work~\cite{GrotovRi16} on online L2R which relates it to contextual bandit formulation~\cite{LangfordZh08}. However, to the best of our knowledge, all the published algorithms on online L2R assume the ability to inject random noise in the results for exploration (e.g. when using Thompson Sampling). In our deployment, we do not have the liberty to do this since our users see only one result and are quite sensitive to the quality of our results.

Our work is also related to lexicalized approaches to search which learn a large number of features explicitly based on query terms~\cite{BWGCSQCW10}. Many recent neural approaches to IR also take into account terms using their embeddings~\cite{GyselRiKa18}~\cite{ChNaDo17}. Also our work is closely related to a plethora of question answering work in NLP~\cite{ChFiWeBo17}, some of which use paraphrases~\cite{FaZeEt13} for question answering which is akin to our notion of query similarity. However, there is some evidence~\cite{GuoFaAiCr16} to suggest that retrieval should be treated differently from question answering. ~\cite{BoSaBaZa15} adapt the notion of query similarity towards semantically equivalent questions rather than actual paraphrases. One of the works that comes close to us is~\cite{RadlinskiJo05} who create chains of related queries from web logs and extract lexical features from queries. Reader should also refer to~\cite{AACDHR19} for an analysis of using conversational platforms for search.
The key features that makes our KB SaaS setup and algorithm apart from the related work are that we 1) use online learning since each organization uses \spoke differently, 2) do not inject noise in our predictions, and 3) allow for the possibility of KB to be changed during a deployment.

\vspace{-0.1in}
\section{Experiments and Metrics}\label{sec:experiments}
In this section, we will present experiments using our online L2R approach. We will present datasets, baselines and compare our online L2R with competitive baselines. We will show how our online L2R strategy with adaptive learning outperforms competitive baselines.

\begin{table}[t]
\scriptsize
\begin{tabular}{|l|c|c|c|c|}
\hline
Client Id & domain(s) & \#KB & \#q & avg q per KB \\\hline
1 & HR, IT, Safety & 10  & 13  & 1.4  \\\hline 
2 & IT, HR, Finance, Office  & 135  & 192  & 1.43  \\\hline 
3 & HR & 122  & 285  & 2.40  \\\hline 
4 & HR, Design, Facilities & 58  & 185  & 3.21  \\\hline 
5 & HR, Office& 202  & 384  & 1.96 \\\hline 
6 & IT  & 61  & 235  & 3.87  \\\hline 
7 & HR, Marketing, IT, Office & 30  & 62  & 2.07 \\\hline 
8 & Business Ops, Legal, HR & 206  & 1352  & 6.57  \\\hline 
9 & \makecell{IT, HR, Ops,  Product, \\Customer Support} & 39  & 100  & 2.62 \\\hline 
10 & \makecell{Marketing, Sales, \\ Product, Data Analysis} & 38  & 51  & 1.42  \\\hline 
11 & Knowledge & 25  & 56  & 2.24 \\\hline 
12 & \makecell{Legal, Product, HR, Ops\\ IT, Education, Engineering} & 102  & 385  & 3.78 \\\hline 
\end{tabular}
\caption{\label{tab:dataset} Details of client datasets showing details of the datasets and the diversity of domains in our data.}
\normalsize
\end{table}

\begin{table*}[h]
\begin{center}
\footnotesize
\begin{tabular}{|l?p{17pt}|p{17pt}|p{17pt}|p{17pt}?p{17pt}|p{17pt}|p{17pt}|p{17pt}?p{17pt}|p{17pt}|p{17pt}|p{17pt}?p{22pt}|}
\hline
ClientId & \multicolumn{4}{c|}{BM25} &  \multicolumn{4}{c|}{Static L2R (artificial + client)} &  \multicolumn{4}{c|}{Our Algorithm (static + adapt)} & $\Delta$F1 \% \\\hline 
 & Prec&Rec&F1 & MRR& Prec&Rec&F1 &MRR &Prec&Rec& F1 &MRR&   \\\hline
1 & 0.714	& 0.714	& 0.714 & 0.773 & 
1 &  0.857 & 0.923 & 0.939 & 1 & 0.857 & 0.923  & 0.939 & 0 \\\hline 

2  & 0.37 &	0.4 &	0.384 & 0.567 & 
0.719 & 0.644 & 0.679 & 0.864 & 0.809 & 0.764 & 0.786 & 0.9 & \textbf{15.8} \\\hline 

3 & 0.627 & 0.639	& 0.633 & 0.708 & 
0.852 & 0.765 & 0.806 & 0.896 & 0.901 & 0.806 & 0.851 & 0.915 & \textbf{5.6} \\\hline 

4 & 0.563	& 0.658 & 0.607 & 0.765 & 
 0.582 & 0.763 & 0.66 & 0.906 & 0.681 & 0.774 & 0.725 & 0.936 & \textbf{9.8} \\\hline 

5  & 0.649 &	0.68	& 0.664 & 0.708 & 
0.762 & 0.73 & 0.746 & 0.856 & 0.804 & 0.74 & 0.771 & 0.868 & \textbf{3.4} \\\hline 

6 & 0.357 &	0.559	& 0.436 & 0.596 & 
0.297 & 0.646 & 0.407 & 0.807 & 0.467 & 0.744 & 0.574 & 0.886 & \textbf{41} \\\hline 

7 & 0.556	& 0.714	& 0.625 & 0.801 & 
0.605 & 0.778 & 0.681 & 0.877 & 0.636 & 0.778 & 0.7 & 0.87 & \textbf{2.8} \\\hline 

8 & 0.372	& 0.388 & 0.38 & 0.562 &  
 0.608 & 0.547 & 0.576 & 0.753 & 0.738 & 0.71 & 0.724 & 0.851 & \textbf{25.7} \\\hline 

9 & 0.618 & 0.667	& 0.642 &0.802 & 
0.798 & 0.775 & 0.786 & 0.948 & 0.83 & 0.765 & 0.796 & 0.948 & 1.3 \\\hline 

10  & 0.515	& 0.625 &	0.565 & 0.763& 
0.494 & 0.714 & 0.584 & 0.92 & 0.575 & 0.75 & 0.651 & 0.923 & \textbf{11.5} \\\hline 

11 & 0.862 &	0.893 &	0.877 & 0.887 & 
0.944 & 0.911 & 0.927 & 0.971 & 0.962 & 0.893 & 0.926 & 0.971 & -0.1\\\hline 

12  & 0.731	& 0.74	& 0.735 & 0.839 & 
0.897 & 0.631 & 0.741 & 0.91 & 0.917 & 0.714 & 0.803 & 0.939 & \textbf{8.4}\\\hline
\Xhline{2\arrayrulewidth}

Average  & 0.578 & 0.640 & 0.605 &  0.731 & 
0.713 & 0.730 & 0.710 & 0.887 &
0.777 &	0.775 & 0.769 &	0.912 &
\textbf{10.43}\\\hline 

\end{tabular}
\end{center}
\caption{\label{tab:results}\footnotesize Details of results comparing our online L2R approach with BM25 and static training. We compare precision, recall, F1, and MRR. $\Delta$ F1 contains relative F1 point improvements of our online L2R approach over static training baseline. Static L2R vastly outperforms BM25 (17.5\% relative improvement in average F1.) Our online algorithm provides massive relative F1 improvements over the static one --- 10.4\% on average and upto 41\%. We also improve average MRR by 2.8\%.}
\normalsize
\end{table*}

\subsection{Datasets for Offline Training and Evaluation} \label{sec:dataset}
We use two kind of datasets for our experiments.
\paragraph{Artificial dataset:} To train the parameters for static pairwise match model $\statics$, we (the authors of this paper) created a dataset containing 364 questions matched to 83  KB articles. For each question, there is a single unique KB answer (similar to as shown in Table~\ref{kb-examples}).
\ignore{
The KB and questions are selected from 12 common workplace domains: \texttt{IT, HR, Engineering, Legal, Design, Security, Finance, Sales, Recruiting,  Facilities, Events, Office Admin}. 
}

\paragraph{Client dataset:} We obtain data from real world feedback from 12 of our clients. These clients were chosen due to the high level of product engagement and the diversity of use cases they cover. Each data set contains a stream of events generated as a result of users from that organization naturally interacting with an older version of our system. Each event in the stream is timestamped and has one of the following types:
\begin{itemize}[itemsep=0pt,topsep=3pt,leftmargin=*]
\item KB creation or updation: a KB article is created or updated.
\item KB deletion: a KB article is deleted.
\item Query search and feedback: tuple $(q, d, u, y)$ where a user $u$ searches with a query $q$; our system responds with an article $d$ and the user gives a feedback $y \in \{+, -\}$.
\item Expert feedback: tuple $(q, d, e, +)$ for cases our system could not answer query $q$, and a domain expert $e$ responds with article $d$. 
\end{itemize}
We discard all tuples with negative feedback for offline training and evaluation as they do not provide the ground truth. Table~\ref{tab:dataset} shows information and statistics of our client datasets.

\begin{table}[t]
\small
\begin{center}
\begin{tabular}{|l|r|}
\hline
Feature & Description \\\hline
\textit{LemmaComparison} &  Compare lemma in query and text\\\hline 
\textit{Term Match} & Unigram and bigram dot product  \\\hline 
\textit{Synonym Match} & Term overlap with synonyms \\\hline 
\textit{Word2Vec match} & IDF-weighted word2vec dot product \\\hline 
\textit{Acronym match} & Query and text acronym overlap\\\hline 
\end{tabular}
\end{center}
\caption{\label{tab:features}Feature templates for static scoring $\statics$.}\vspace{-0.1in}
\normalsize
\end{table}

\subsection{Training and Evaluation}\label{sec:trainandeval}
\paragraph{Training:} We train the offline model $\statics$ only on the artificial dataset. We trained a LambdaMart model~\cite{Burges10} as well as a simple linear RankSVM model~\cite{Joachims02} minimizing pairwise ranking loss~\cite{Burges10} and found no performance difference. So for simplicity, we use the linear model for $\statics$. We fine tune the hyperparameters --- $\beta$ (weight of positive queries), $\gamma$ (weight of negative queries), $\delta_e$ (weight for expert feedback), and $\delta_u$ (weight update for user feedback), and $\tau$ (the score threshold) --- on development data from four of our clients, maximizing the total micro F1. We exclude the development data from evaluation.

\paragraph{Evaluation:} We use the client dataset for evaluation. For each query, there is at most 1 correct answer. For each dataset, we simulate real user and expert feedback. We run Algorithm~\ref{al:overview} going over the event stream in the order of timestamps. We provide negative or positive feedback using ground truth. When the system makes a mistake, we reveal the correct KB article only when the event corresponds to expert feedback as that is the role of experts (Sec.~\ref{sec:system}). We evaluate all search algorithms using four metrics: \textit{Precision@1}, \textit{Recall@1}, \textit{F1@1}, and \textit{MRR} (mean reciprocal rank.)
\subsection{Experimental Comparisons} \label{sec:comparison}
\paragraph{Baselines:} We compare our online L2R algorithm with two strong baselines: the BM25 algorithm~\cite{RWJHK96}, that is the de facto standard available in publicly available search software like Apache Solr\footnote{http://lucense.apache.org/solr}, and the static pairwise match baseline (only $\statics$) trained with RankSVM~\cite{Joachims02}. For fair comparison, we train the baseline $\statics$ model on the artificial data as well as  the development data used for tuning our algorithm (Sec.~\ref{sec:trainandeval}). Comparing with the $\statics$-only baseline shows how much we can gain from doing online learning from user feedback. 

\paragraph{Features for $\statics$:} We use the match feature templates in Table~\ref{tab:features} for the $\statics$ model. For term match we use modified term frequency reprsentations to normalize for document length~\cite{SinghalBuMi96}. For Synonyms, we use the PPDB dataset~\cite{GaBeCa13}. For word vectors, we use GLOVE embeddings~\cite{MSCCD13}. We compute these features with the query and four textual items obtained from KB: \textsc{title}, \textsc{body}, \textsc{keywords}, and \textsc{all} (\textsc{title}, \textsc{body}, and \textsc{keywords} concatenated).

\paragraph{Our algorithm setting:} For our algorithm, we set query similarity kernel $s_{qsim}$ to be the cosine similarity over unigrams and bigrams. We use  $g=\sumtopk$ with $k=5$ as a query similarity aggregator.

\paragraph{Results:}
Table~\ref{tab:results} shows the performance of online learning, static learning, and BM25. Our algorithm outperforms the static baseline by average 10.4\% relative improvement in F1 (and up to 41\%), which in turn vastly outperforms BM25 (17.5\% relative improvement in average F1). Notably the Pearson Correlation coefficient between $\Delta$F1  in Table~\ref{tab:results} and average query per KB is 0.66. This hints that incorporating adaptive learning is more likely to help for cases with higher queries per KB article.

\vspace{-0.1in}
\section{Conclusion and Future Work}
We presented a production system for supporting internal KB search inside organizations that gets continually better at responding to queries using conversational feedback. We leverage online learning and incorporate various practical concerns into the design of our algorithm and scoring function. In the future, we aim to inject neural networks-based features into our online learning setup by using deep learning-based query similarity functions.

\small
\bibliographystyle{named}
\bibliography{naaclhlt2019}

\end{document}